\pdfoutput=1
\RequirePackage{fix-cm}

\documentclass[twocolumn]{svjour3}
\usepackage{multicol}
\usepackage{graphicx}
\usepackage[square,numbers]{natbib}
\usepackage{bm}
\usepackage{subfig}
\usepackage[T1]{fontenc}
\usepackage{booktabs}
\usepackage{mathptmx}
\usepackage{soulutf8}
\usepackage{url}

\usepackage[colorinlistoftodos]{todonotes}
\setuptodonotes{inline}

\begin{document}

\title{Automatic travel pattern extraction from visa page stamps using CNN models
\thanks{This project has received funding from the European Union’s Horizon 2020 research and innovation programme under grant agreement No 833704}
}

\titlerunning{Automatic Passport Visa Page Stamps Recognition}

\author{Eimantas Ledinauskas  \and
        Julius Ruseckas       \and
        Julius Marozas        \and
        Kasparas Karlauskas   \and
        Justas Terentjevas    \and
        Augustas Mačijauskas  \and
        Alfonsas Juršėnas
}

\institute{E.~Ledinauskas \and
           J.~Ruseckas \and
           J.~Marozas \and
           K.~Karlauskas \and
           J.~Terentjevas \and
           A.~Mačijauskas \and
           A.~Juršėnas \at
           Baltic Institute of Advanced Technology,
           Pilies 16-8, LT-01403, Vilnius, Lithuania\\
           \email{eimantas.ledinauskas@bpti.eu}
}

\date{Received: date / Accepted: date}

\maketitle

\begin{abstract}
Manual travel pattern inference from visa page stamps is a time consuming activity and constitutes an important bottleneck in the efficiency of traveler inspection at border crossings. Despite efforts to digitize and record the border crossing information into databases, travel pattern inference from stamps will remain a problem until every country in the world is incorporated into such a unified system. This could take decades.
We propose an automated document analysis system
that processes scanned visa pages and automatically extracts the travel pattern from detected stamps. The system processes the page via the following pipeline:
stamp detection in the visa page;
general stamp country and entry/exit recognition;
Schengen area stamp country and entry/exit recognition;
Schengen area stamp date extraction.
For each stage of the proposed pipeline we construct neural network models and train then on a mixture of real and synthetic data. We integrated Schengen area stamp detection and date, country, entry/exit recognition models together with a graphical user interface into a prototype of an automatic travel pattern extraction tool. We find that by combining simple neural network models into our proposed pipeline a useful tool can be created which can speed up the travel pattern extraction significantly.
\keywords{Stamp recogniton \and Date recognition \and Automated document analysis \and Neural networks}
\end{abstract}

\section{Introduction}
\label{intro}

There is a need to increase the speed and accuracy of document analysis at all lines of international border control \cite{kulju2018framework}. Improved document checking should improve travelers convenience and security. Advances in computer vision could enable a new level of automation in this field.
Therefore automating the work of border guards is an area of growing interest. The are methods proposed for, e.g., document anonymization \protect{\cite{bouma_document_2020}}, passport scanning, facial and fingerprint recognition \cite{oostveen_automated_2014, gorodnichy_art_2015}.

Border guards have to infer the travel pattern of a person by looking at stamps and dates in visa pages of the passport. They also have to validate the stamps for counterfeits and infer if there is an anomaly of the travel pattern. This is important for early prevention of international criminal activity. Often this analysis cannot be done appropriately due to long queues and requirement of quick processing of people entering the country. There are efforts to create the unified international databases where border crossing information could be registered (e.g. EU Entry-Exit System
\footnote{https://ec.europa.eu/home-affairs/policies/schengen-borders-and-visa/smart-borders/entry-exit-system\_en}). Automatic rule-based analysis of this database would solve the issues mentioned above. However, complete disappearance of visa page stamps does not seem plausible in the near future as only a subset of countries are engaging in these unified database projects.

\begin{figure*}
    \centering
    \includegraphics[width=\textwidth]{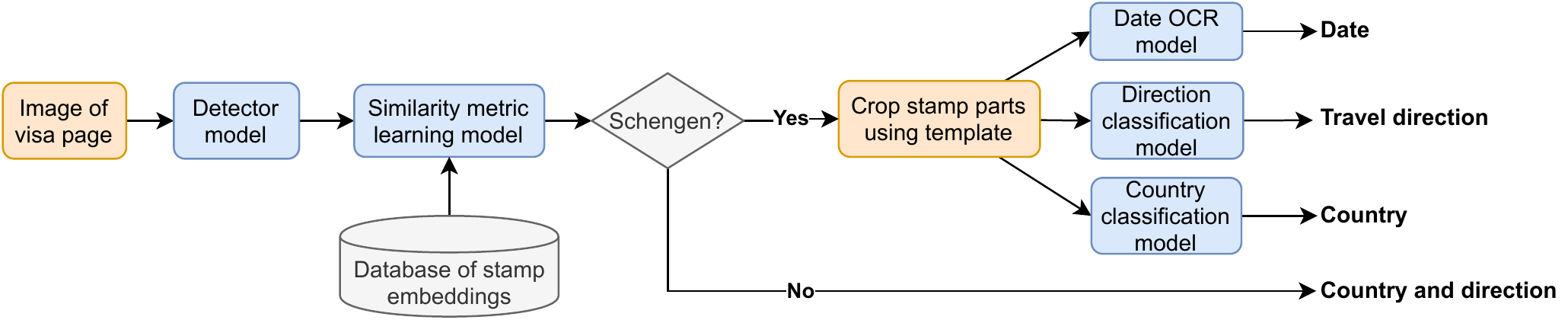}
    \caption{The proposed page processing pipeline.
    The image of visa page could be obtained by a passport scanner or from a video feed by the passport facing camera that is mounted on the table.
    After that the stamps are detected with a detector model.
    The detected stamps are later compared with a set of known stamp embeddings by using the similarity metric learning model which provides the country and travel direction of the most similar stamp in the database of stamp embeddings.
    If the stamp belongs to a Schengen country, then additionaly the Schengen template and several additional models are being used to extract date, travel direction and a country.
    }
    \label{fig:pipeline}
\end{figure*}

In order to automate the travel pattern extraction from stamps in visa pages, we propose a document analysis system that processes the scanned page via the following pipeline:
stamp detection in the visa page;
general stamp country and entry/exit recognition;
Schengen area stamp country and entry/exit recognition;
Schengen area stamp date extraction.
The page processing stages are shown in figure~\ref{fig:pipeline}.
For each stage of the proposed pipeline we construct neural network models.
If the detected stamp belongs to the Schengen area, a separate pipeline for Schengen area stamps is selected, as shown in figure~\ref{fig:pipeline}. We choose this because dedicated Schengen area stamp models achieve better performance than the general models. For general stamp classification we employ a similarity learning model that searches for a most similar stamp in the database. In contrast, for Schengen area stamps we crop parts of the stamp using a predefined template and subsequently classify those parts. We find this approach to be robust enough because Schengen area stamps have a standardized layout across all of the member countries. An additional stamp segmentation stage can be used before stamp similarity model to remove overlapping stamps.

We find that in the case of Schengen area stamps the automatic travel pattern extraction can be made precise enough for practical applications even with modest amounts of training data. In the case of countries outside of the Schengen area this problem is a lot harder due to variability of stamp formats. However, we think that our proposed approaches could still be used in practical applications if significantly more training data would become available.


\subsection{Related work}
\label{related}

Currently there are three widely employed methods of document analysis: 1) manual data entry and processing, 2) methods based on template extraction, 3) template-less machine learning methods.

Template-based document analysis systems \cite{Cesarini1998,Rusinol2013,Schuster2013,dAndecy2018} locate the required text by utilizing the distance and direction from surrounding keywords. They require an initial setup of hard-coded rules for every template. However, such methods often fail when a document with unseen template is encountered \cite{Rusinol2013}. To improve template-based methods, in \cite{Dhakal2019} an one-shot template-matching algorithm invariant to changes in position is proposed. Methods that work on unseen document formats were proposed in \cite{Palm2017,Holt2018}. In \cite{Palm2017} the CloudScan system that employs a recurrent neural network (RNN) is presented.

Stamp detection in documents requires separate methods. For stamp detection various features based on color, shape and textual content can be employed. In \cite{Zhu2006,Roy2011} localization is performed using boundary shape. Segmentation based on color-related features is used in \cite{Micenkov2011,Dey2015}, whereas in \cite{Nandedkar2015,Singh2018} textual information is used. In \cite{ahmed_generic_2013} where the authors used statistical techniques based on shape, color and size to separate stamp from its surroundings. For a review of similar methods see \cite{Alaei2016}. A pipeline for stamp detection and classification has been proposed in \cite{Zaaboub2020}. The steps involved are edge detection, texture based segmentation and final classification using feed-forward neural network. However, the proposed method is applicable only to isolated stamps.

A more sophisticated approach implementing neural network (YOLOv2) to detect stamps was suggested in \cite{smolinski_segmentation_2020}. Furthermore, a great success was achieved in \cite{ronneberger_u-net_2015} using U-Net to segment biological structures and since both cells and stamps feature small details we could naturally expect accurate segmentation results on stamps.

Stamp recognition (country, entry/exit, etc.) problem can be reduced to matching an image of a new stamp against a database of known stamps. Feature engineering based on shape or color has been used in an attempt to recognize stamps used in various purposes \cite{kacprzyk_efficient_2011,Forczmanski2012,Forczmanski2013,Petej2013, forczmanski_stamps_2015}.
However, such methods are limited and could hardly be employed in environments that require high accuracy like border control.

More advanced approaches used Siamese networks to learn a similarity measure for face recognition \cite{taigman_deepface_2014} which proved to drastically improve performance. Similarly, \cite{wang_csrs_2019} have used Siamese networks for Chinese seal recognition which in its essence is very similar to our task of recognizing stamps from visa pages. Like in our work, they also faced the difficulty of having scarce training data and solved it by using synthetic data generation.

Most of above mentioned articles implement only a part of the tasks needed to solve the travel pattern recognition from visa pages:
stamp detection, segmentation, classification, optical character recognition.

To the best of our knowledge there is a lack of published work that demonstrates a pipeline for automatic travel pattern extraction based on purely deep learning approaches.
The goal of this work is to demonstrate such a pipeline.
Based on deep learning success in other fields there is a reason to believe that this approach will outperform hand-crafted methods.

Finally, the success of a neural network hugely depends on the variety and quality of a training data set. However, in our case we had to work with very limited resources thus artificial data generation was essential. In \cite{su_deep_2017} authors suggested a solution to this problem by first augmenting the ground truth image and then pasting it on different backgrounds, achieving significantly higher accuracy than on real images alone. The major drawback is that ground truth images (without a background) are required.

\section{Methods}
\label{methods}
\subsection{Data preparation}
\label{data_preparation_and_augmentation}

We have created datasets for model training using stamp images from openly accessible sources together with stamp images provided by Lithuanian state border guard service.
Since the number of images obtained in this way is too small, we used synthetic images that were created by superimposing stamp images on a randomly selected background image (sampled uniformly). For synthetic image creation segmented stamp images are required. The segmentation task was automated using a stamp segmentation model because manual stamp segmentation is very time consuming.

For stamp segmentation a model described in subsection~\ref{methods_segmentation} has been employed. The segmentation model was trained on manually segmented stamps, subsequently the trai\-ned model has been used to segment stamps from 1822 images. The automatically segmented images along with the manually segmented stamps were used to generate data for the Siamese network described in subsection~\ref{sec:similarity_methods}.

\begin{figure*}
    \centering
    \includegraphics[width=0.9\textwidth]{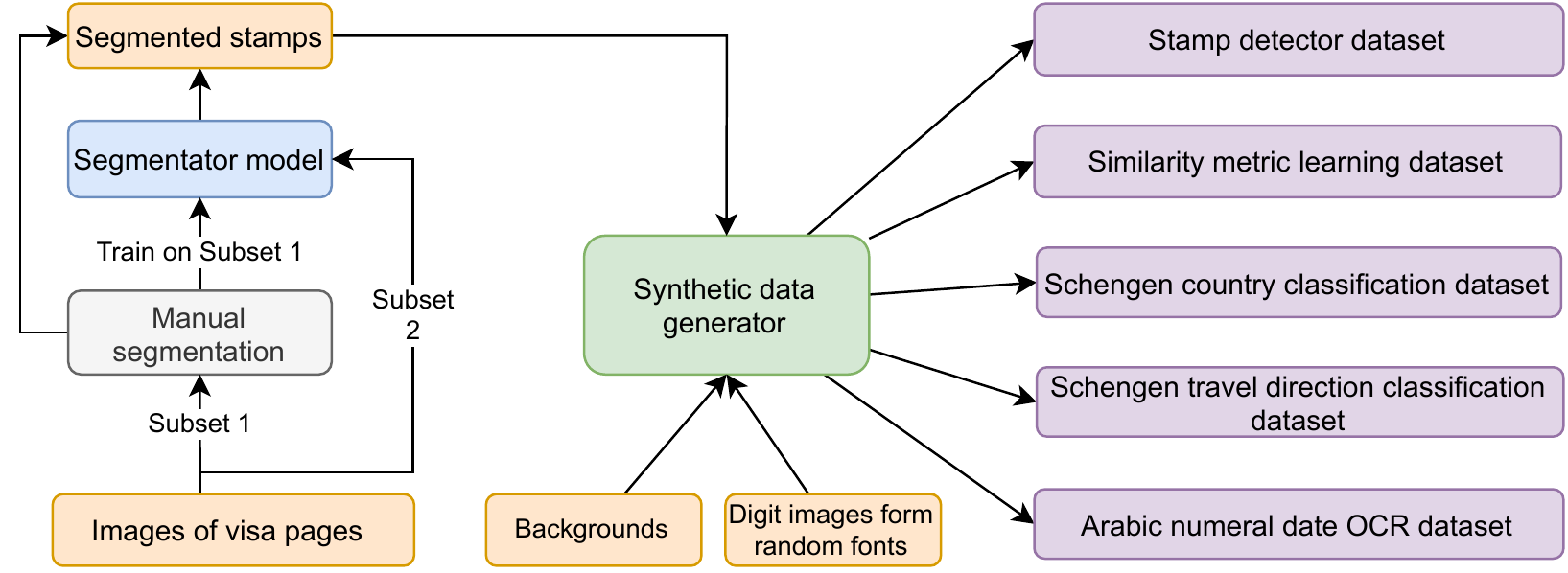}
    \caption{
        Pipeline for generation of model training datasets.
        A subset of images of stamps in visa pages are being segmented manually while another subset of these images are segmented by the segmentation model.
    }
    \label{fig:datasets_pipeline}
\end{figure*}

The full pipeline for generation of model training datasets is shown in figure~\ref{fig:datasets_pipeline}. In addition to synthetic visa page images used in stamp detection and classification training, we crop direction and country symbols from Schengen area stamps to create datasets for country and direction classification models. Digit images cropped from Schengen area stamps as well as digit images from random fonts are used to create the dataset for date recognition.
The detailed data preparation for each model is described in corresponding subsection presenting the model (subsections~\ref{sec:methods_stamp_detection}, \ref{methods_segmentation}, \ref{sec:similarity_methods}, \ref{sec:methods_country_entry_exit}, \ref{sec:methods_date_recognition}).

\subsection{Model for detecting affinely transformed stamps}
\label{sec:methods_stamp_detection}

The architecture of the stamp detection model used in this work is mainly inspired by the single-shot detector YOLO \cite{Redmon2016_YOLO}. The main difference is that the model outputs not the bounding boxes but bounding quadrangles, similarly to EAST text detection model \cite{Zhou2107_EAST}. Quadrangles are represented by four independent corner points and thus have more degrees of freedom than typical rectangular bounding boxes. ResNet18 \cite{He2016} is used as a network backbone to extract the feature maps. The feature maps are then sent through two 1x1 convolutional layers in parallel which reduce the number of channels to 1 and 8. The first feature map is interpreted as a grid of confidence scores (after acting on it with sigmoid) and the second map is interpreted as a grid of vectors of bounding quadrangle coordinates. This architecture is simple but works surprisingly well on rectangular stamps where the quadrangle coordinates simply signify the corners of the stamp. We chose to use quadrangles instead of the usual vertical boxes because sometimes the stamps are rotated (even up to $180^\circ$) and they must be unrotated for date optical character recognition (OCR) model to work successfully. We also experimented with predicting the standard bounding box together with rotation angle but found that at least for stamp detection the approach with quadrangles works significantly better.

Segmented stamps (both manually and with model described in sec \ref{methods_segmentation}) were used to generate synthetic training data by inserting them on various backgrounds. The size, insertion location and rotation angle of stamps were all chosen randomly from uniform distributions. The size varied from 30\% to 70\% of the background image width and it was fixed between stamps in the same  image so that every stamp in the same example would be of the same size (as stamp size does not vary in reality). The rotation angle varied from -180$^\circ$ to 180$^\circ$. The number of stamps inserted in a single image varied from 1 to 3 with equal probabilities and stamps were allowed to overlap with each other. The background images before the insertion of stamps were augmented by random horizontal and vertical flipping and affine transformation with rotation angle varying from -90$^\circ$ to 90$^\circ$ and shear angle from -16$^\circ$ to 16$^\circ$. After insertion the final image was augmented with blur (Gaussian, average or median chosen randomly), additive Gaussian noise, pixel and coarse (from 3\% to 0.15\% of image width) dropout, intensity addition (per channel from -10 to 10), hue and saturation addition (from -20 to 20) and intensity multiplication (per channel from 0.5 to 1.5). All of the parameters were sampled from uniform distributions.

The loss function was calculated as a sum (with equal weights) of crossentropy loss on confidence scores and mean square error loss on quadrangle coordinates.

For optimization the ADAM \cite{Kingma2015_ADAM} algorithm was used with running average parameters $\beta_1 = 0.9$ and $\beta_2 = 0.999$. The learning rate was set to $10^{-3}$ at the start and then reduced to $10^{-4}$ and $10^{-5}$ at the epoch numbers 35 and 80 respectively. Training was done for 90 epochs with 5000 synthetic images per epoch and batch size of 32.

\subsection{Stamp segmentation model}
\label{methods_segmentation}

Training data for the segmentation model was generated from a set of 299 manually segmented stamp images and 80 background images. The stamp images were combined with background images, as described in subsection \ref{data_preparation_and_augmentation}.
Each stamp was randomly rotated by up to $5^{\circ}$ in either direction. The background image was randomly flipped (horizontally and/or vertically, with probabilities of 0.5 each), rotated by up to $10^{\circ}$ in either direction and had its contrast and brightness adjusted. If the background image was too small, it was upscaled to four times the dimensions of the stamp image. A random part of the background image was picked to put behind the stamp. Finally, an additional stamp was randomly placed in one of the four pre-set locations in the corners of the image, undergoing all of the same transformations the original stamp did.

The architecture used for segmentation model was a modified U-Net \cite{ronneberger_u-net_2015} with same padded convolutions and batch normalization \cite{batchnorm}. The weights of the convolutional layers were initialized using Kaiming initialization \cite{kaiming_init}.

The binary cross-entropy loss was used in model training. For optimization ADAMW \cite{adamw} algorithm has been employed, with weight decay of 0.01. The model was trained for a total of 30 epochs using a 1 cycle learning rate schedule \cite{smith_disciplined_2018} with cosine annealing, the maximum learning rate was 0.001. The total training set size was 7231.

To evaluate the performance of the model, mean Dice coefficient was computed for every batch. The Dice coefficient of the n-th sample image is defined as
\begin{equation}
    \mathcal{D} =2\left(\sum_{p=1}^{P}y_p\cdot\sigma(x_p)\right)
    \left(\sum_{p=1}^{P}[y_p + \sigma(x_p)]\right)^{-1}\,,
\end{equation}
where $y_p$, is the actual class of the pixel and $\sigma(x_p)$ is the predicted class of the pixel for each of the $P$ pixels of the n-th sample image. Mean Dice coefficient was not used for model optimization. The best validation metrics were achieved in epoch 23.

\subsection{Similarity metric learning for stamp recognition}
\label{sec:similarity_methods}

For general stamp country and entry/exit recognition we propose to use Siamese networks.
Siamese networks \cite{taigman_deepface_2014} are able to work with multiple input images simultaneously and select relevant features for various machine learning tasks, including the estimation of similarity. Similarity estimation can be used for classification where instead of classifying inputs into predefined classes, Siamese networks find the most similar example in the database of class examples. The model consists of two parallel embedding networks that share weights between themselves. A similarity score can then be assigned to pairs of images by measuring the Euclidean distance (or some other distance metric) between their corresponding embedding vectors produced by the networks. While this approach requires more bookkeeping because of the database (in comparison to the classification model), the architecture is able to perform one-shot learning, i.e., work with a dynamically changing class set which is important in the context of automatic border control as it would not be convenient to retrain the network whenever new classes were added to the database. Also classification model requires hundreds or even thousands of examples for each different stamp type while similarity learning model can be trained with as little as 5-10 examples per class (on condition that there are many different classes).

The goal of learning a similarity metric is to train a model that would work in such a way that the distance between embeddings of two images of same stamp class would be as small as possible and the distance between embeddings of two different stamp classes would be as large as possible (note that same class here and below refers to stamps being from the same country and having the same entry or exit direction). We achieved this by employing a Siamese network that was trained using triplet loss, similarly to \cite{schroff_facenet_2015}.

The architecture of our embedding network is a modified ResNet18 \cite{He2016,he_bag_2018}. One of the network modifications is the introduction of a learnable scalar $\gamma$ (initially set to 0) that parameterizes the strength of the shortcut connection (SkipInit \cite{De2020_skipinit}; similar parameter has also been used in \cite{zhang_self-attention_2019}):
\begin{equation}
  \bm{y}_i = \gamma \bm{x}_i + \bm{r}_i\,.
\end{equation}
Here $\bm{y}_i$, $\bm{x}_i$, and $\bm{r}_i$ are the output, shortcut and residual vectors, respectively.

ResNet18 is used to construct feature maps which are converted to embedding vectors by the head (the last layers) of the network. Our network head consists of average pooling layer, followed by fully connected layer and finally a layer that normalizes the output to unit vectors. We normalize the output so that we could more accurately measure how similar output embeddings are to each other and also to be able to choose an appropriate margin for triplet loss. The input to the network has the size $128 \times 128 \times 3$, with the third dimension being the RGB color channels, and the output is a
16384-dimensional ($2^{14}$) vector. All the input images are first resized so that the largest value of width and height become 128 pixels and then padded by black pixels to get a square of the desired size. To increase generalization a random subset of up to three image augmentations are applied during training. The set of possible augmentations includes of rotations, Gaussian blur, Gaussian noise, per channel dropout, changes in contrast, brightness, hue and saturation.

Triplet margin loss incentivizes the model to cluster visa pass stamps by their class. The loss function takes a triplet which consists of three embeddings: one arbitrary embedding called an \textit{anchor}, a \textit{positive} embedding which is of the same class as the anchor, and a \textit{negative} embedding from a different class than the anchor. The minimum desired distance between anchor-positive and anchor-negative embedding pairs is parameterized by a hyperparameter $\alpha$ called a margin. We use the standard Euclidean distance to measure distance. Therefore, the triplet margin loss is given by the equation
\begin{equation}
   \mathcal{L}(a, p, n) = \mathrm{max}\{0, d(a, p) - d(a, n) + \alpha\}\,,
\end{equation}
where anchor, positive, and negative inputs are represented as $a$, $p$, and $n$ respectively;
\begin{equation}
  d(x, y) = \Vert x - y \Vert_2^2
\end{equation}
is the square of the Euclidean distance.

As the model learns, anchors and positives get closer together while anchors and negatives become more distant. In each batch, we made use of the hard triplet mining method \cite{schroff_facenet_2015}: we first computed embeddings and then took each item as an anchor and generated triplets by matching it with the hardest positive (the one furthest away from the anchor) and the hardest negative (the one closest to the anchor) from the current batch. We chose this method because we found that the naive random triplet generation lead to a state where the loss of the majority of the triplets was close to zero and hard cases that were infrequent, although not that rare, had too little of an effect and substantially slowed down learning.

\begin{figure*}
\centering
\begin{multicols}{2}
    \includegraphics[scale=0.4]{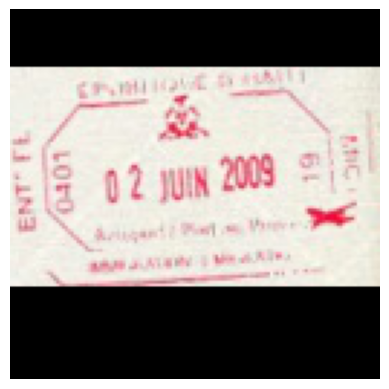}\par
    \includegraphics[scale=0.4]{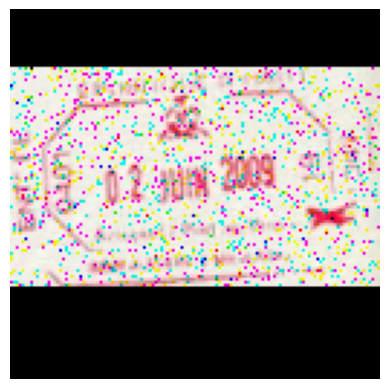}
\end{multicols}
\begin{multicols}{2}
    \includegraphics[scale=0.4]{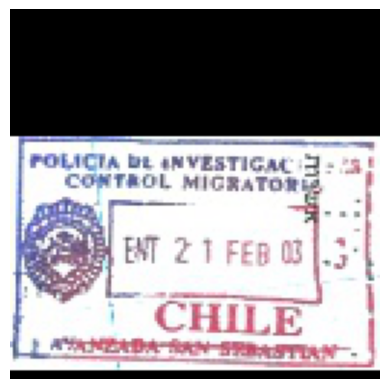}\par
    \includegraphics[scale=0.4]{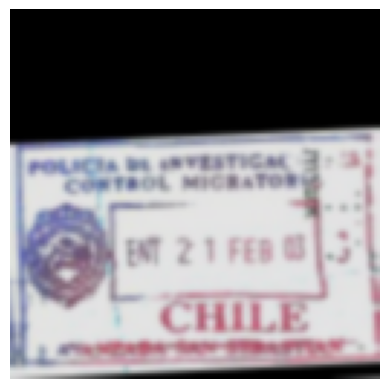}
\end{multicols}
\begin{multicols}{2}
    Input images padded to square\par
    Respective augmented versions
\end{multicols}
\caption{Samples of stamp images provided as an input to the similarity metric model.}
\label{sample_images}
\end{figure*}

We used the 40050 images generated as discussed in subsection~\ref{data_preparation_and_augmentation} to train and evaluate our models. A total of 2121 segmented stamp images, making up 267 distinct country-direction classes (this number is not divisible by 2 because for some countries we did not have examples of both directions), were superimposed over 80 background images, as in subsection~\ref{methods_segmentation}. Up to three additional stamps, each number equally likely, were randomly placed in the corners of the image (with a probability of 0.8), undergoing all of the same transformations the original stamp did. The process was repeated until each class had 150 of distinct training images, 40050 images in total. Proportions of the image subsets, used to generate the splits are shown in table~\ref{tab:stamp_bg_split}.

\begin{table}[!ht]
    \centering
    \caption{Proportion of image subsets used to generate training and test data.}
    \begin{tabular}{@{}lll@{}}
        \toprule
        Subset & Total & Test ratio \\ \midrule
        Stamps & 2121 & 0.10 \\
        Backgrounds & 80 & 0.14 \\ \bottomrule
    \end{tabular}
    \label{tab:stamp_bg_split}
\end{table}

Examples of synthetic images can be found in figure~\ref{sample_images} where input images padded to a square are displayed on the left and augmented versions of those images are shown on the right. There were 267 classes in total with 150 images for each class. We put aside 27 ($\approx 10\%$) of these classes to what we call \textit{unseen validation} set as images of those classes were generated from a separate set of stamps and backgrounds that are never shown to the model during training. We then took another $10\%$ of images for validation purposes and trained on the remaining $80\%$ of the images. To calculate the accuracy on different datasets after training, we also took 2 images from each class per dataset to a separate database, i.e., the training and the validation databases had 240 classes with 2 images per class which is a total of 480 images, while the unseen validation database had 54 images (27 classes $\times$ 2 images per class).

Apart from accuracy, we also tracked a difference metric which measured the average size of differences between the anchor-positive and anchor-negative databases. Therefore, as our model trained, the difference metric was used as a complement to loss to reflect how it improved and got closer to convergence.

For model training we used the ADAMW optimizer with 0.01 weight decay and one-cycle learning rate schedule \cite{smith_disciplined_2018}
with a maximum learning rate of 0.01. Our batch size was set to 128 which was the maximum that fit in the VRAM of the computers of the service providers that we used. However, it should be noted that it could be worthwhile to increase the batch size as this could benefit the hard triplet generation process. We used a margin of $\alpha=0.75$ in the triplet loss formula. The model was trained for 40 epochs before it converged.

\subsection{Schengen stamp country and entry/exit classifier}
\label{sec:methods_country_entry_exit}

Schengen area stamps have a standardized layout in all of the member countries. This can be used to greatly simplify the recognition of country and crossing direction symbols and make it more reliable compared to similarity learning approach described in the previous section. The subimages of country code and entry/exit symbols can be easily segmented after the detection of the bounding quadrangle. Then the symbols can be recognized directly with separate specialized models. With this in mind, we chose to implement the country and direction recognition from these subimages with simple convolutional neural network classifiers. The country symbols could be recognized with OCR model but in this case the number of countries in Schengen area is only 26 so direct classification is a more simple and reliable approach.

In both cases the same convolutional neural network architecture was used with 5 convolutional layers (kernel size $3 \times 3$, stride 2) followed by a global average pooling and a fully connected layer to map the resulting feature vector into the class confidence scores. Batch normalization was done after every convolutional layer. The number of channels after the first convolution was 16 and was doubled after every following convolution. The input image size was set to $64 \times 64$ for country recognition and to $32 \times 32$ for entry/exit recognition.

Training data for the classifiers was created as follows: stamp images were generated as described in section~\ref{sec:methods_stamp_detection}; subsequently the stamps were cropped to the regions where the country code or entry/exit signs are located. The training procedure of the neural networks was similar to the procedure described in section~\ref{sec:methods_stamp_detection}. Training was done for 60 epochs (1000 images per epoch) using batch of size 64. The starting learning rate $10^{-3}$ was reduced to $10^{-4}$ at epoch 40.

\subsection{OCR model for dates in Schengen stamps}
\label{sec:methods_date_recognition}

In the case of Schengen area stamps the date format is standardized, with fixed number of digits and their location. Therefore the problem of automatic date recognition in Schengen stamps is greatly simplified. Similarly to the case of country and entry/exit symbols (section~\ref{sec:methods_country_entry_exit}), the image of the date can be easily cropped from the stamp image after the detection of the stamp bounding quadrangle.

With these simplifications in mind, we created a simple specialized neural network architecture for date recognition. First, the date image is passed through ResNet18 to construct a feature map. This feature map is flattened and passed through two fully connected layers that have 400 and 60 neurons, respectively. The final output of size 60 is interpreted $10 \times 6$ matrix describing class confidence scores for 6 date digits. Due to the fully connected layers, the input image size must always be of fixed size ($512 \times 128$ pixels in our case) and the network assumes that there are always six date digits in the image. We also tried more flexible OCR algorithms but found that this specialized architecture significantly outperforms them.

\begin{figure}
    \centering
    \includegraphics[width=0.8\linewidth]{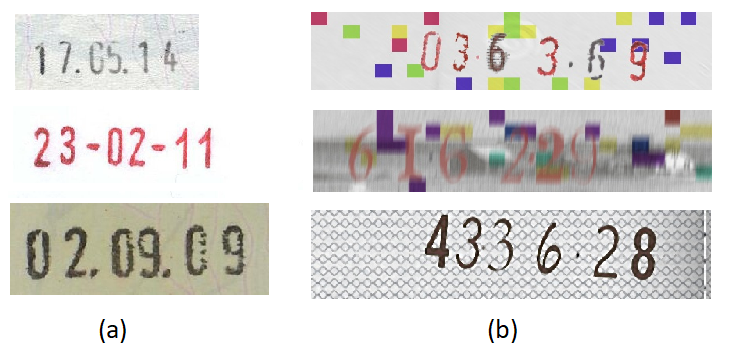}
    \caption{Date reading OCR dataset examples: (a) real test examples, (b) synthetic training examples.}
    \label{fig:ocr_dataset_examples}
\end{figure}

The model was trained with synthetic data. Synthetic dataset of date images was generated by superimposing individual images of digits onto random backgrounds (sampled uniformly). The digit images have been selected from the following sets:
1) digit images segmented from the images of real stamps (134 examples);
2) digit images generated by random fonts (2566 examples, sampled uniformly).
Date image was generated from a random sequence of digits (sampled uniformly) superimposed onto a random crop of a background (in total 88 backgrounds were used). Spacing between digits, date aspect ratios, and rotation angle of the date was also randomly changed at each example date. Resulting image was further augmented using imgaug augmentation library \cite{imgaug} with the following augmentations: motion blur, coarse dropout, additive Gaussian noise, add to hue, add to saturation, channel shuffle. Dates have been fixed to the following format: "XX-XX-XX" or "XX.XX.XX" where X denotes any single digit. In total 16 000 training and  2000 validation synthetic date examples were generated.
For testing 171 real stamp date images have been used. Samples of real and synthetic images are shown in figure~\ref{fig:ocr_dataset_examples}.

The training procedure of the neural network was similar to the procedure described in section~\ref{sec:methods_stamp_detection}. Training was done for 20 epochs (19342 images per epoch) using batch of size 50. The starting learning rate $10^{-3}$ was reduced to $10^{-4}$ at epoch 15.

\section{Results}

\subsection{Stamp detection}

\begin{figure}[tbp]
  \centering
  \includegraphics[width=\linewidth]{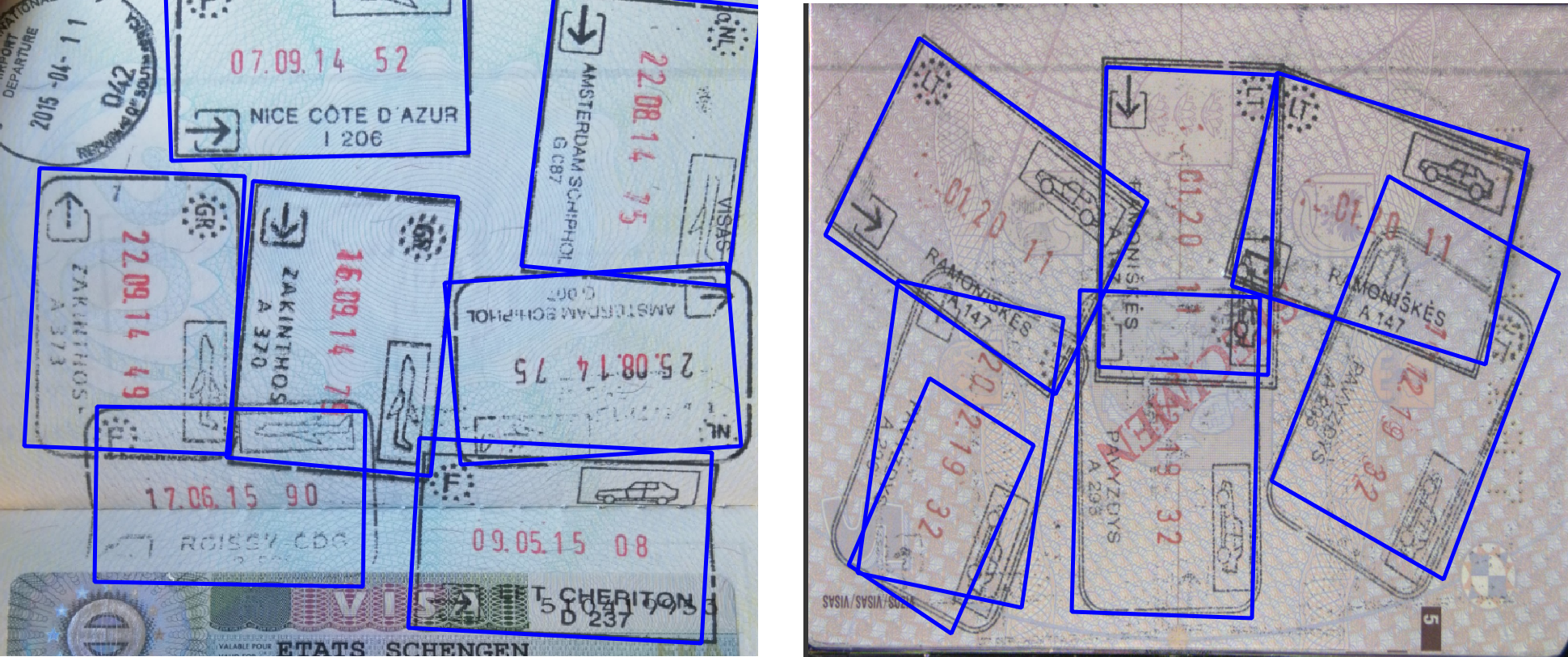}
  \caption{Schengen area stamp detection examples. The predicted bounding quadrangles are shown in blue.}
  \label{fig:stamp_detection_examples}
\end{figure}

After training the Schengen area stamp detection model reaches 99.1\% precision, 87.8\% recall and 0.78 intersection over union (IoU) on the synthetic validation dataset (which was generated by using the unseen backgrounds and unseen stamps). It is very hard to reach recall higher than ~90\% on this dataset as augmentations used are quite strong and significant fraction of stamps strongly overlap while also being strongly rotated relative to each other.

As most of the real data were used for generating training and validation datasets, we did not calculate detection metrics on the real testing data as we did not have a lot of examples left. But visually the model works really well and detects the stamps with almost perfect accuracy when there is no overlap. Several detection examples are shown in figure~\ref{fig:stamp_detection_examples} and it can be seen that even in the case of significant overlaps and rotations the model most of the time detects the stamp and also its rotation correctly. We also find that the detection quadrangles most of the time are precise enough in order to use them for country and direction symbol cropping which are needed for later processing steps (see sec. \ref{sec:methods_country_entry_exit}).

Since a relatively small neural network is used for stamp detection, the detection step is fast. This model is able to process 17 frames per second on a laptop with Intel Core i7-8665U CPU. Thus the performance of the model is sufficient for real-time detection of stamps in a video feed even without using graphical processor acceleration.

\subsection{Schengen stamp country and entry/exit recognition}

\begin{figure}[tbp]
  \centering
  \includegraphics[width=0.8\linewidth]{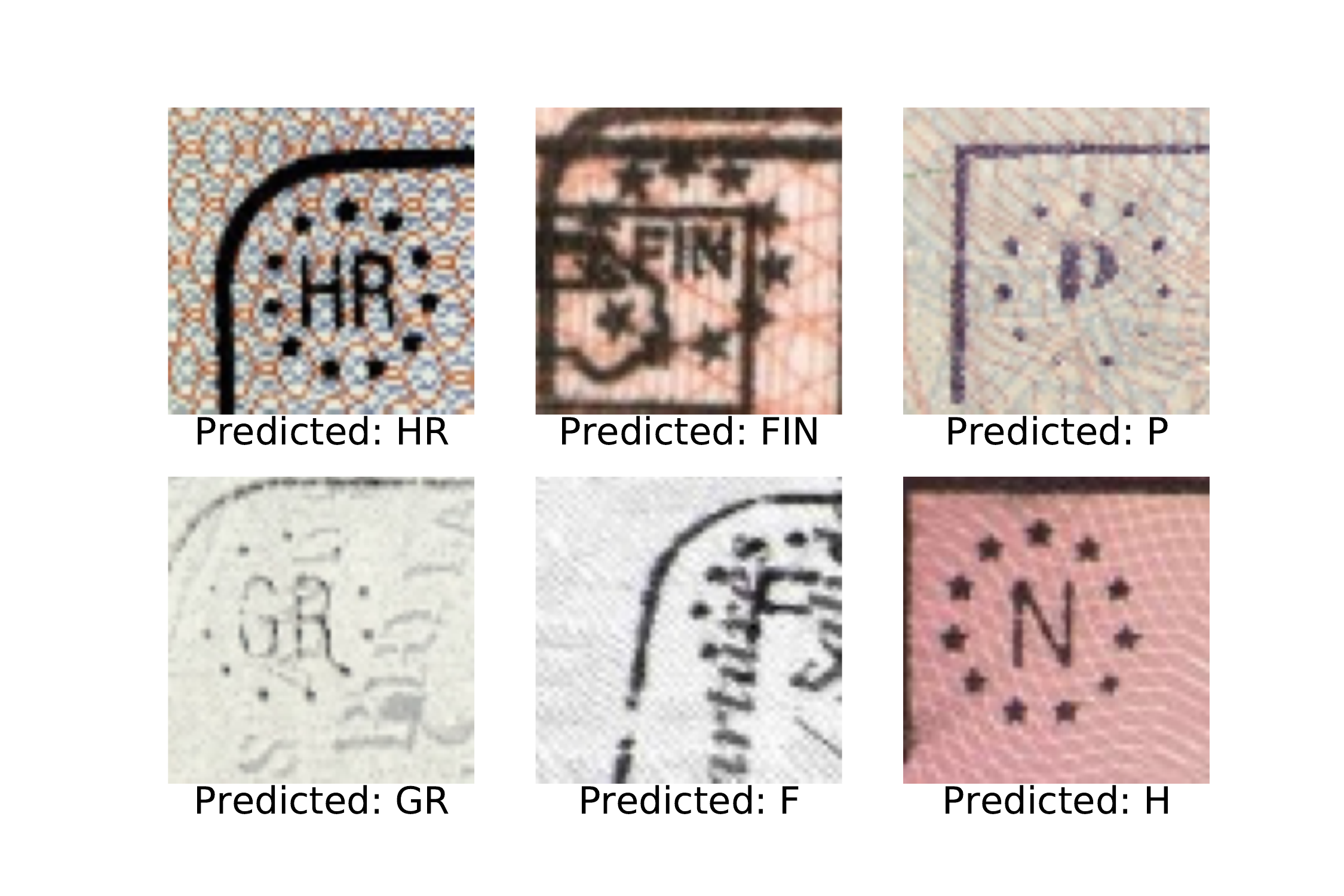}
  \caption{Country recognition examples.}
  \label{fig:country_recognition_examples}
\end{figure}

\begin{figure}[tbp]
  \centering
  \includegraphics[width=0.8\linewidth]{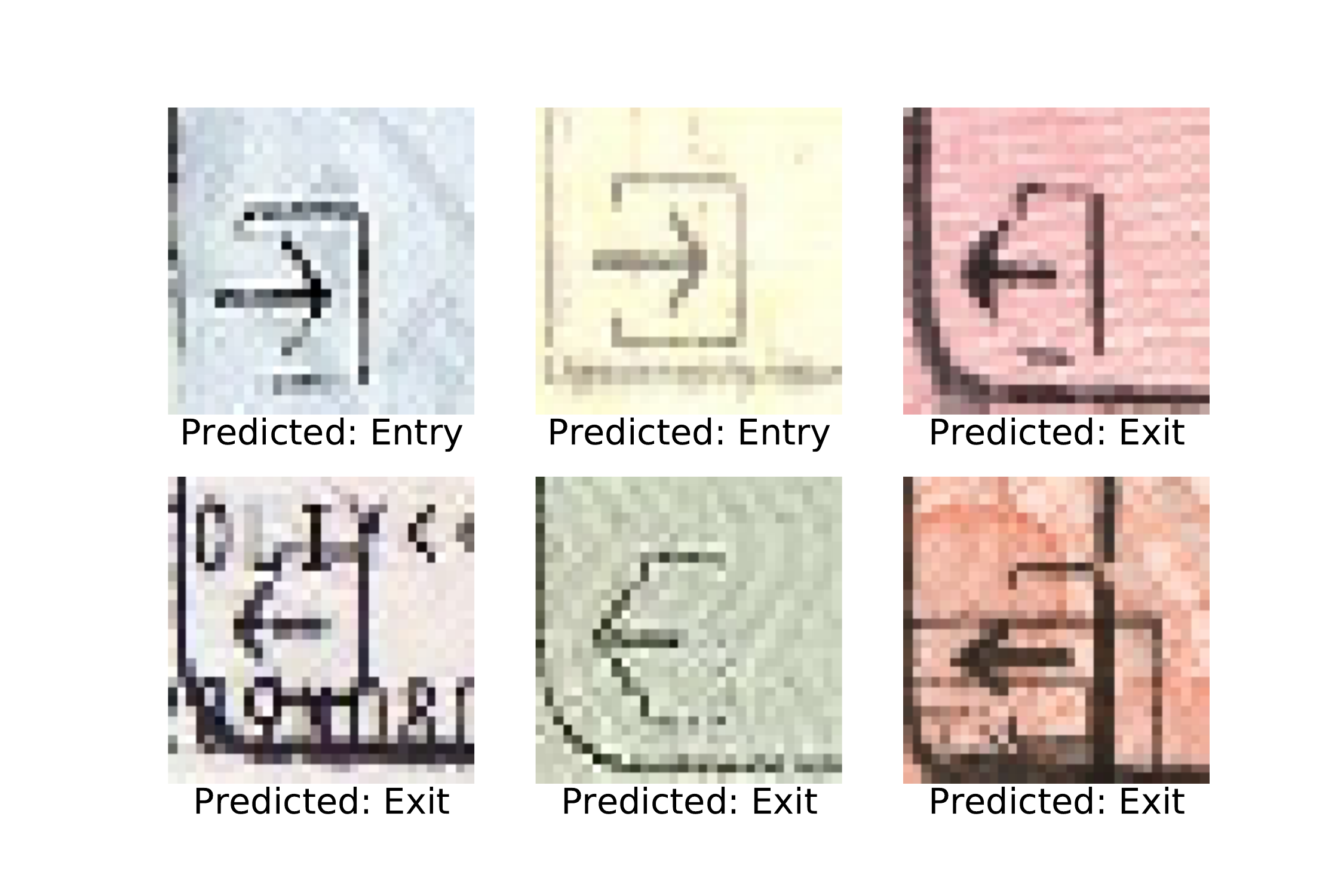}
  \caption{Entry/exit recognition examples.}
  \label{fig:inout_examples}
\end{figure}

After training the country and Entry/Exit classification models reach 96.4\% and 99.3\% validation accuracies respectively. Some examples of input images and predictions can be seen at figures \ref{fig:country_recognition_examples} and \ref{fig:inout_examples}. Some examples are hard because of overlapping stamps or other interfering patterns in the background.

When testing both models on real test examples we observed that for both models the accuracies decline significantly. This might be due to small number of training examples used and different data distributions between testing and training/validation datasets. Still, mistakes happen on minority of examples and these models can be used in practice to hasten the human work. It is highly probable that increasing the amount of training data would drastically improve the generalization.

\subsection{Shengen stamp date recognition}

\begin{figure}[tbp]
  \centering
  \includegraphics[width=\columnwidth]{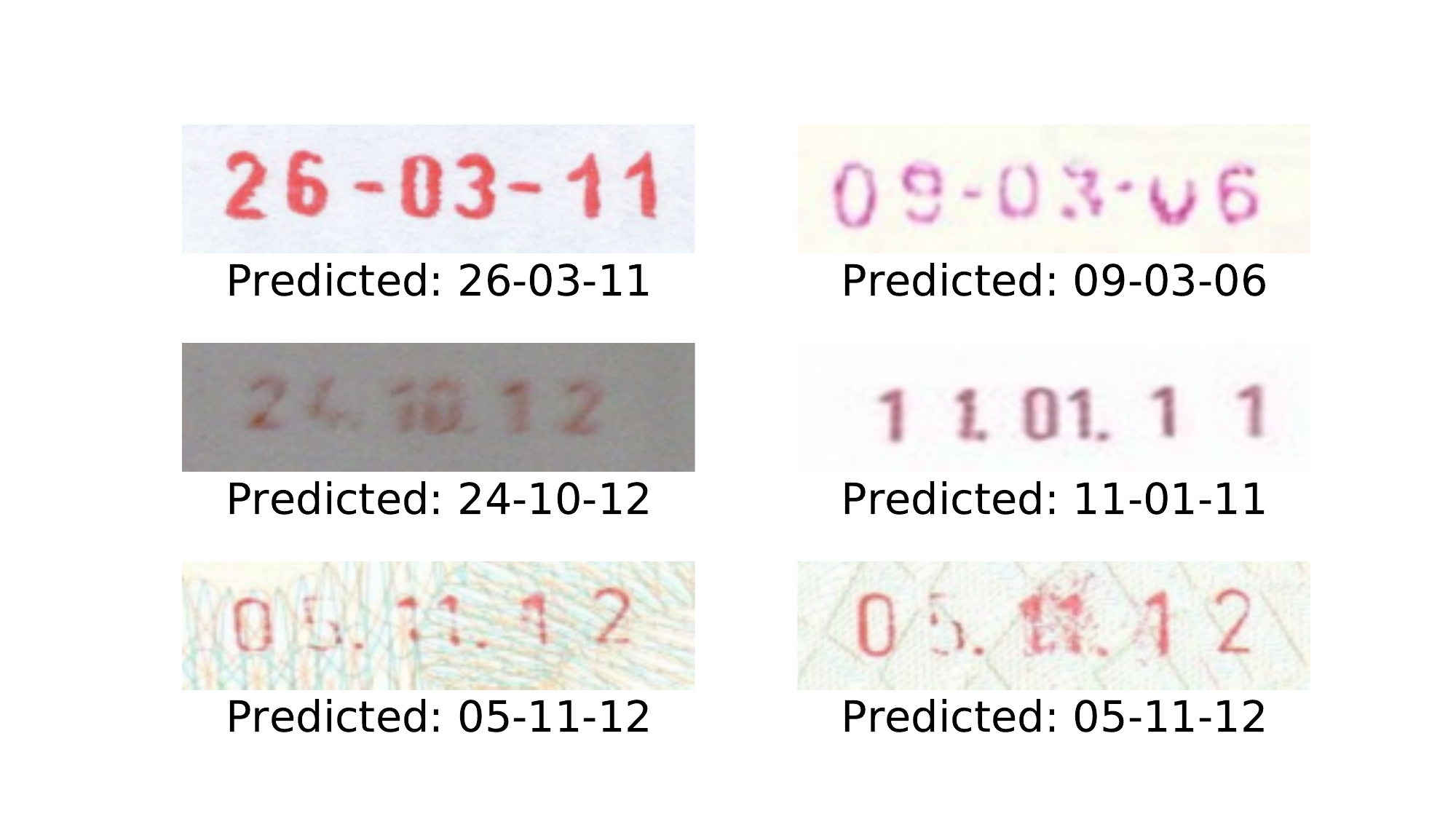}
  \caption{Date recognition examples.}
  \label{fig:date_recognition_examples}
\end{figure}

After training the date recognition model reached 99.5\% character accuracy on validation data. It also generalized to the real testing data really well and displayed almost perfect accuracy in our final tool testing scenarios. Moreover, errors that occur in most cases are related to wrong stamp bounding quadrangles  provided by detection model. Several examples of images and model predictions are shown in figure~\ref{fig:date_recognition_examples}. As can be seen in these examples, model recognizes digits accurately even in the cases of blurry images, incomplete characters and problematic backgrounds.

\subsection{Stamp segmentation}

\begin{figure*}
    \centering
    \includegraphics[width=\textwidth]{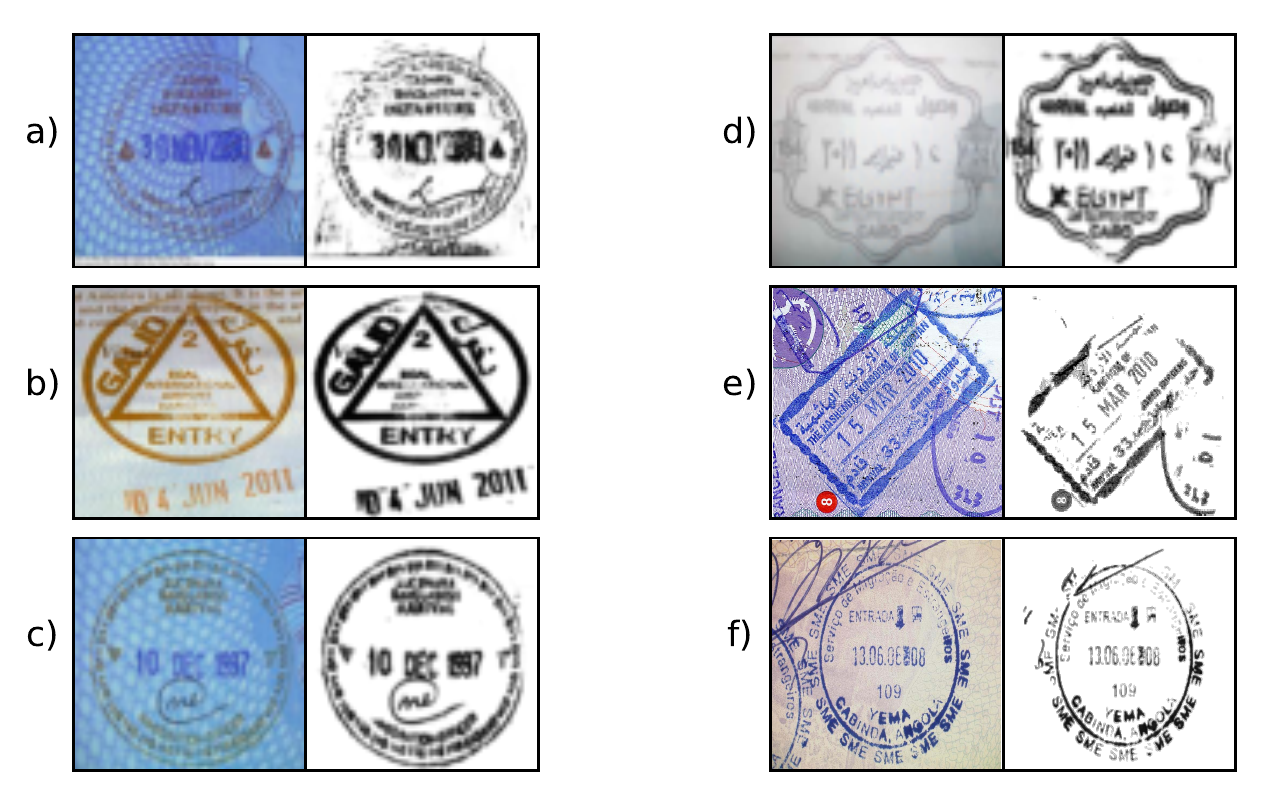}
    \caption{Pairs of inputs from the validation dataset and segmentation model outputs.}
    \label{fig:unet_examples}
\end{figure*}

After training the stamp segmentation model reached the mean Dice coefficient $\mathcal{D} = 0.909$ on the validation set. The model successfully segmented 1822 stamp images that were used as training data for other models.
An example of the performance of the segmentation model on unseen (synthetic) data is shown in figure~\ref{fig:unet_examples}. Stamp a) demonstrates problems with segmentation in highly figured backgrounds with darker features present. A portion of the highly figured background is misidentified as a part of the stamp. Stamps b) and c) demonstrate the model’s ability to accurately segment stamps from slightly figured backgrounds (i.e. patterns or overlapping text). Stamp d) shows the model’s performance on clearer backgrounds. Finally, model’s performance on overlapping stamps is shown in fig. 9 e) and f). It can be seen that the model can successfully separate the main stamp given that the secondary stamp takes up relatively small part of the image. However, if two stamps are similar in size accuracy drops as both stamps are recognized as one.

\subsection{Similarity metric learning for stamp recognition}

\begin{table}
\centering
\caption{Accuracy of country recognition by similarity model}
\label{accuracy_country_dir}
\begin{tabular}{@{}llll@{}}
\toprule
 & Train & Validation & Unseen validation \\
\midrule
Percentage & 97.51\% & 98.25\% & \textbf{92.87\%} \\
Count & 30658/31440 & 3537/3600 & \textbf{3711/3996} \\
\bottomrule
\end{tabular}
\end{table}

\begin{figure}
    \includegraphics[width=\columnwidth]{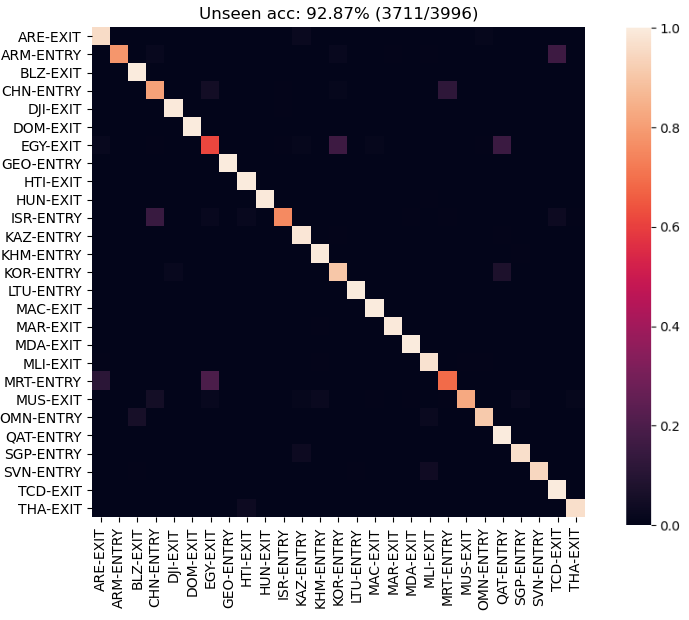}
    \caption{Confusion matrix on the unseen countries dataset}
    \label{confusion_matrix}
\end{figure}

The detailed performance of the model after training is summarized in
table~\ref{accuracy_country_dir}. The contents of the datasets used for training and validation are discussed in subsection~\ref{sec:similarity_methods}.
The confusion matrix is shown in figure~\ref{confusion_matrix}. As you can see, the model achieves satisfying accuracy, the confusion matrix is close to an identity matrix.

\begin{figure*}
    \centering
    \begin{multicols}{2}
        \includegraphics[scale=0.29]{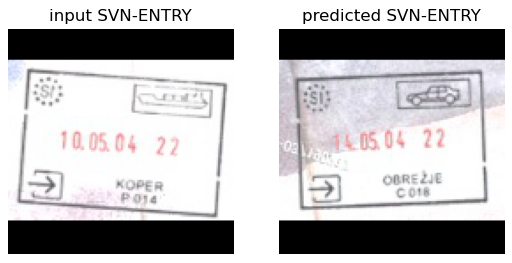}\par
        \includegraphics[scale=0.29]{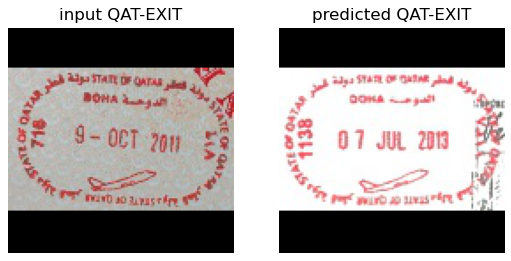}\par
    \end{multicols}
    Correct (easy)\par
    \begin{multicols}{2}
        \includegraphics[scale=0.4]{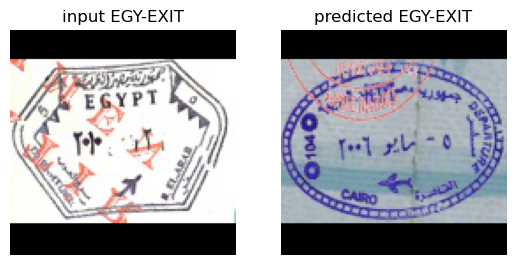}\par
        \includegraphics[scale=0.29]{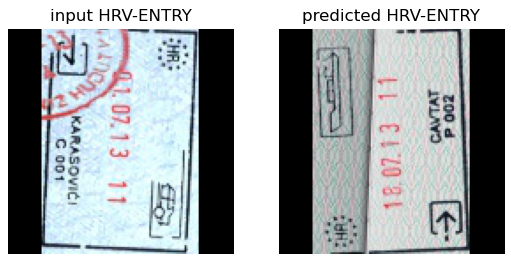}\par
    \end{multicols}
    Correct (hard)\par
    \begin{multicols}{2}
        \includegraphics[scale=0.4]{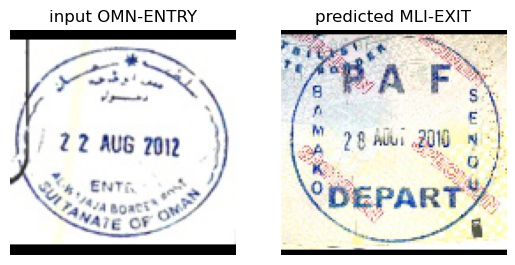}\par
        \includegraphics[scale=0.4]{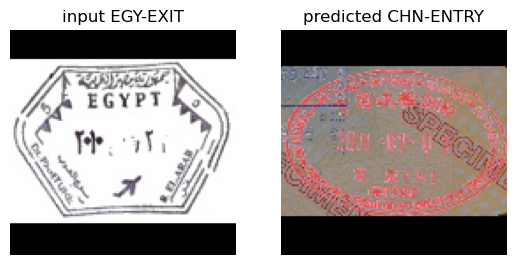}\par
    \end{multicols}
    Wrong (both from unseen)
    \caption{Samples of input-output image pairs}
    \label{samples_input_output}
\end{figure*}

First two rows of figure~\ref{samples_input_output} include some examples of input-output pairs that our model got right. It can be seen that the model is able to correctly match a wide range of inputs, some of which would be extremely hard or impossible even for humans. It should also be noted that the model exhibits invariance of many features that are often hard to train for, e.g.\ leftover artifacts, shape, color, rotation, squeezing, blur, bad quality.

Referring to the third row of figure~\ref{samples_input_output}, it can be seen that the model is still not perfect when it comes to matching inputs. For one, errors come from the model mixing up classes that have stamps of similar shape and color. Also, there are some inputs and database items that are of very poor quality, so it might happen that the model incorrectly matches them because of the lack of features to predict on. Finally, there were errors where it is hard to tell why the model failed (e.g.\ no similarity in shapes or colors) which could be further investigated in future work.

\begin{table}
\centering
\caption{Accuracy of the model getting the country right, but not necessarily the direction}
\label{accuracy_country}
\begin{tabular}{@{}llll@{}}
\toprule
 & Train & Validation & Unseen Validation \\
\midrule
Percentage & 99.58\% & 99.83\% & 92.87\% \\
Count & 31307/31440 & 3594/3600 & 3711/3996 \\
\bottomrule
\end{tabular}
\end{table}

\begin{table}
\centering
\caption{Rate of misclassification due to wrong detection of direction}
\label{accuracy_dir}
\begin{tabular}{@{}llll@{}}
\toprule
& Train & Validation & Unseen validation \\
\midrule
Percentage & 82.99\% & 90.48\% & N/A \\
Count & 649/782 & 57/63 & N/A \\
\bottomrule
\end{tabular}
\end{table}

\begin{figure*}
    \centering
    \begin{multicols}{2}
        \includegraphics[scale=0.4]{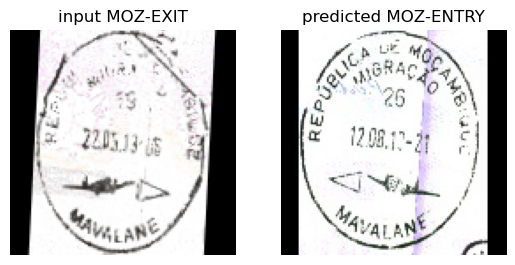}\par
        \includegraphics[scale=0.4]{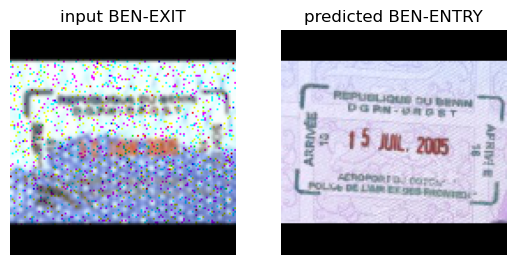}\par
    \end{multicols}
    \begin{multicols}{2}
        \includegraphics[scale=0.4]{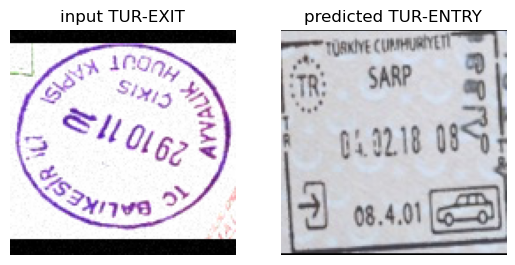}\par
        \includegraphics[scale=0.4]{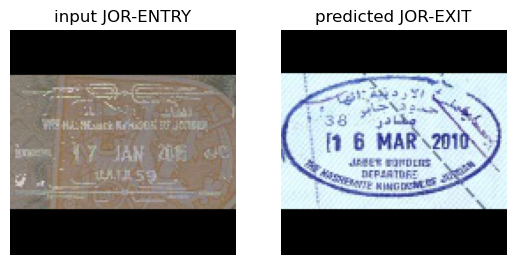}\par
    \end{multicols}
    \caption{Samples of wrong direction input-output image pairs}
    \label{samples_input_output_direction}
\end{figure*}

As can be seen from table~\ref{accuracy_country}, the trained model gets the country right with over 99\% accuracy on the validation dataset. Note that the accuracy of unseen validation dataset is the same as in table~\ref{accuracy_country_dir} because there were no classes that would refer to the same country but different direction in this dataset. Table~\ref{accuracy_dir} suggests that a very high percentage of our errors come from the model getting the country right but mixing up the directions. Some examples such input-output pairs can be seen in figure~\ref{samples_input_output_direction}. Evidently, the image on the top left shows that some stamps have only a very small feature (e.g. an arrow) that distinguishes the direction and are very similar otherwise which presents a challenge to the model. Also, the images on the top right and bottom right suggest that some errors could arise form the input images being blurry or of bad quality which, again, might make it hard for the model to read off the correct direction. Finally, the image on the bottom left suggests that the the error could potentially come from the database not being representative enough of the input data so that stamps from the same class that have different shapes or colors are mismatched.

\subsection{Final automatic travel pattern extraction solution}
\label{sec:final_solution}

We integrated Schengen area stamp detection and date, country, entry/exit recognition models (see sections \ref{sec:methods_stamp_detection}, \ref{sec:methods_date_recognition} and \ref{sec:methods_country_entry_exit}) together with graphical user interface into an automatic travel pattern extraction tool. For country and direction recognition we chose to use the approach described in sec. \mbox{\ref{sec:methods_country_entry_exit}} instead of the one based on similarity learning (sec. \mbox{\ref{sec:similarity_methods}}) because Schengen area stamps are all very similar in appearance and thus recognition directly from symbols is more reliable. After loading the scanned visa pages the program can be used to automatically detect the stamps and recognize date, country, entry/exit symbols and then fill the travel pattern table. Of course, our described models do not have perfect accuracies and the system accuracy drops even more then they are used sequentially because of cumulative errors. However, we find that in practice the errors are still rare and, as the user only needs to correct the minority of examples, this system still could greatly speed up the workflow. In the case of Schengen area stamps the accuracy and speed of the automatic travel pattern extraction tool is sufficiently high to be used by border guards, since manual correction is very rarely needed.

\begin{figure*}
  \centering
  \includegraphics[width=\textwidth]{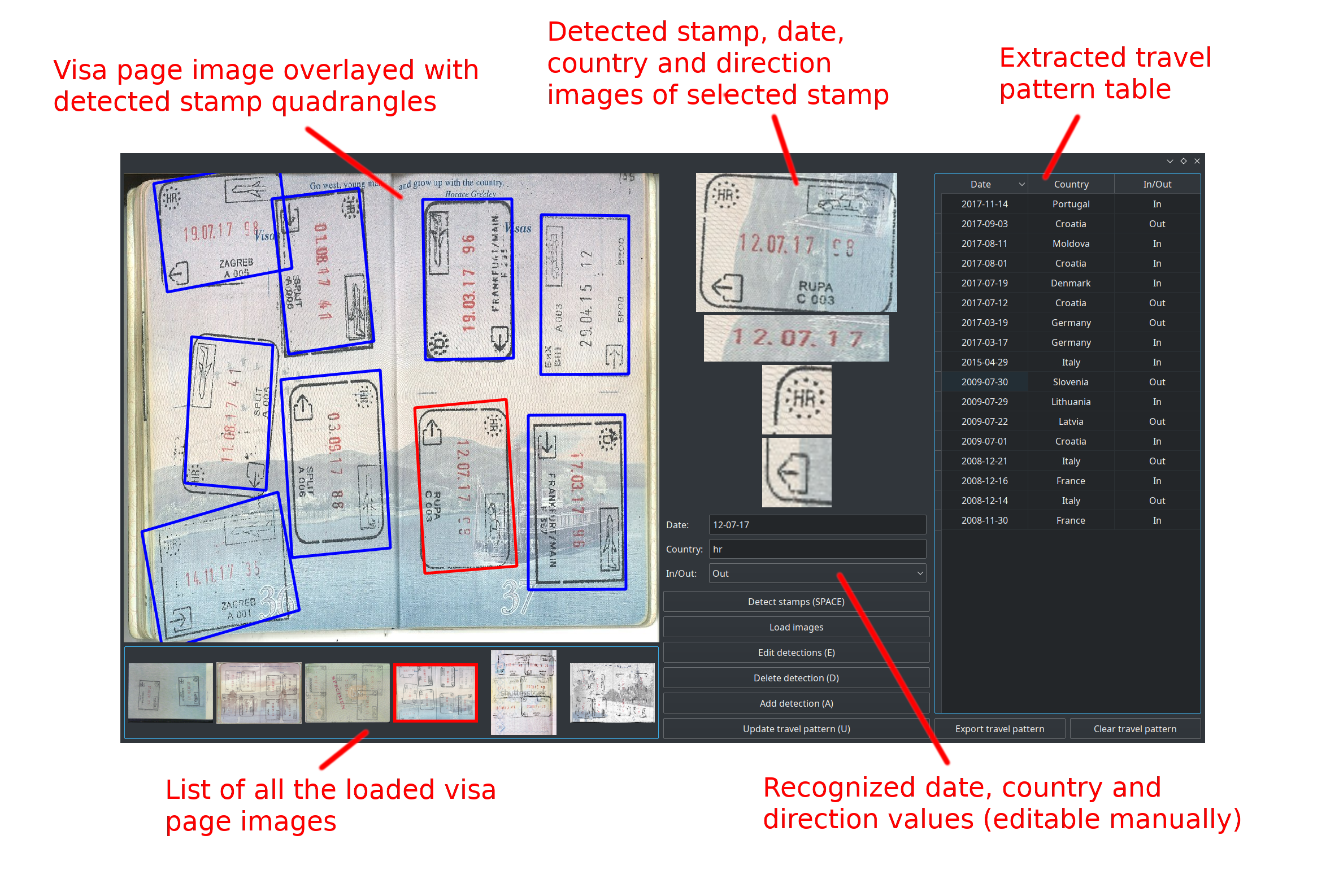}
  \caption{Screenshot of the final travel pattern extraction program.}
  \label{fig:ui_screenshot}
\end{figure*}

Screenshot of the travel pattern extraction program is shown in figure~\ref{fig:ui_screenshot}. After scanning of the visa page, the user is presented with the image of the page on which the detected stamps are indicated. The user can select one of those stamps. In the case of the Schengen area stamp, the image of the selected stamp rotated to horizontal orientation, together with images of the date, country and direction areas are shown. The recognized date, country and direction of travel corresponding to the selected stamp are presented below the images. The user can correct those values manually, if the detection is wrong. Finally, a table of the whole extracted travel pattern is shown on the right hand side.

This prototype works only with Schengen area stamps. It could be extended to work with other country stamps by adding the model described in sec. \mbox{\ref{sec:similarity_methods}}. This model could also be used to differentiate between Schengen area and other stamps (by grouping all schengen area stamps into a single class) and then process the former in the same way as it is done now.

\section{Discussion and Conclusions}

In summary, we proposed an automated document analysis system
that processes scanned visa pages and automatically extracts the travel pattern from detected stamps. In the case of Schengen area stamps the automatic travel pattern extraction tool is precise enough for practical applications. For the countries outside of the Schengen area the performance of the models needs to be improved. However, the problem of general stamp recognition is a lot harder due to variability of stamp formats and limited amount of training data. Surely, even in the Schengen area case the system sometimes makes mistakes and thus it could not be used for fully autonomous solutions. However, we still think that it could greatly increase the efficiency because it would do the majority of the work itself and the human supervisor would only need to correct some rare mistakes. These corrections could also be used to continuously actively train the model and the probability of errors would decrease over time.

Below we discuss how the performance of the models could be further improved. One of the ways is to use more and better data for model training.

The synthetic image generation process could be improved so that the stamp recognition model could be trained on data that is more representative of real world examples. This would require to either further increase the accuracy of the segmentation model or manually cut the stamps by hand. However, another possibility would be to fuse these approaches: after training the segmentation model for some time, the produced stamps could be manually revised by selecting those that were segmented the worst and cut them by hand. These stamps could then be fed into the model again, thus encouraging it to learn and perform better on those harder examples.

Some generated images look unnatural because the contrast between the stamp and the background is too low, to the extent that sometimes it is hard to see where the stamp is even to the human eye. To combat this, the contrast of the generated images could be checked and those with poor quality could be filtered out. Alternatively, the contrast between the dominant colors of the stamp and the background could be checked beforehand to skip over those that would result in poor quality images.

We also noticed that the original images of some stamps were in such poor quality that there was no reason to expect segmentation of those images with an acceptable accuracy. So, it might be useful to manually go over the data and remove such images.

It should be noted, however, that any such changes to synthetic image generation should be implemented with care because, due to the nature of neural networks, it is hard to predict how such changes would affect the model. For example, the model might be able to generalize better or converge faster when the generated data is more distorted than real world data. Therefore, after any substantial changes to the model it is crucial to thoroughly analyze the changes in performance.

The errors in stamp similarity model originate mostly from mistakenly recognized travel direction. Thus one of the first things that could be done to improve the model would be to focus more on direction recognition during training. The image on the bottom left of figure~\ref{samples_input_output_direction} suggests that the error could potentially come from the database not being representative enough of the input data so that stamps from the same class that have different shapes or colors are mismatched.

It may also prove to be useful to make the performance evaluation more strict by using fewer images per class in the databases, i.e., use only one image instead of two. However, this should be done with care as there are classes that have stamps that differ drastically from year to year, so excluding them and leaving only one image per class could show a decrease in accuracy even though the real problem would be that the database is not representative enough.

General stamp classification using stamp similarity model can be further improved by employing the stamp segmentation model to remove pixels belonging to overlapping stamps. This can be achieved by inserting a stamp segmentation stage into processing pipeline.

Finally, the most obvious and fruitful way to improve performance of all models would be to get more real world data to evaluate the model on. However, this is usually much harder to implement.

\section*{Conflict of interest}

The authors declare that they have no conflict of interest.

\bibliographystyle{spbasic}

\end{document}